\documentclass[letterpaper, 10 pt, conference]{ieeeconf}
\IEEEoverridecommandlockouts 
\usepackage{caption}
\usepackage{subcaption}

\let\labelindent\relax
\usepackage{hyperref, url, amsmath, amssymb, booktabs, graphicx, bm, color, algorithm, tabularx, enumitem, flushend}
\usepackage[noend]{algpseudocode}
\usepackage{cleveref}

\newcolumntype{Y}{>{\centering\arraybackslash}X}
\usepackage{booktabs}
\title{\LARGE \bf WeHelp: A Shared Autonomy System for Wheelchair Users }

\author{Abulikemu Abuduweili, Alice Wu, Tianhao Wei, Weiye Zhao
\\
Carnegie Mellon University
}

\usepackage{booktabs}
\begin{document}

\maketitle
\thispagestyle{plain}
\pagestyle{plain}

\begin{abstract}
There is a large population of wheelchair users. Most of the wheelchair users need help with daily tasks. However, according to recent reports, their needs are not properly satisfied due to the lack of caregivers. Therefore, in this project, we develop WeHelp, a shared autonomy system aimed for wheelchair users. A robot with a WeHelp system has three modes, following mode, remote control mode and tele-operation mode. In the following mode, the robot follows the wheelchair user automatically via visual tracking. The wheelchair user can ask the robot to follow them from behind, by the left or by the right. When the wheelchair user asks for help, the robot will recognize the command via speech recognition, and then switch to the teleoperation mode or remote control mode. In the teleoperation mode, the wheelchair user takes over the robot with a joy stick and controls the robot to complete some complex tasks for their needs, such as opening doors, moving obstacles on the way, reaching objects on a high shelf or on the low ground, etc. In the remote control mode, a remote assistant takes over the robot and helps the wheelchair user complete some complex tasks for their needs. Our evaluation shows that the pipeline is useful and practical for wheelchair users. Source code and demo of the paper are available at \url{https://github.com/Walleclipse/WeHelp}.
\end{abstract}

\section{Introduction}
\label{sec:intro}

According to the report \cite{market2021north}, in 2016, there were 3.3 million wheelchair users in the U.S., with 1.825 million of those users aged 65 years and above. This number is predicted to grow every year, with an expected 2 million new wheelchair users per year. The growth rate of the North American Wheelchair Market is 7.9\%, with an estimated value of USD 1.7 Billion by 2028. Over 11 million people needed assistance with activities of daily living (ADLs) or instrumental activities of daily living (IADLs) \cite{brault2012americans}.

In recent years, robotics technology has improved our lives every day in a million ways beyond simply making things more convenient \cite{wang2019artificial,abuduweili2019adaptable}.  As robots are defined to minimize the efforts of humans, their technology allows them to provide fully automated functions with the convenience to use them. Assistive robotics can help people who use wheelchairs better handle the day-to-day tasks that might otherwise be too challenging or awkward \cite{brose2010role}. 

According to the report \cite{pousada2018determining}, wheelchair users' needs are not properly satisfied due to the lack of caregivers. Therefore, in
this project, we develop WeHelp, a shared autonomy system aimed at wheelchair users. Since each kind of wheelchair user is unique in his disabilities or difficulties in specific activities, we would not specify a specific population. Our general targeted population would be all wheelchair users requiring additional assistance in daily activities. 
In regards to how to implement the WeHelp system in specific tasks, our setup is: the care recipient presses the switch to enable controlling the home robot by the remote human caregiver. Giving an example of an opening-the-door scenario. First, the robot with the WeHelp system follows the wheelchair user in accompany mode. The wheelchair user could use speech commands to take control of the robot by himself and open the door. If the wheelchair user has difficulty controlling the robot by himself, who can use speech command to activate the remote control, the remote caregiver receives this command and controls the robot by opening the door. This example shows a shared autonomy system that enables the remote human caregiver and a home robot to work together and provide assistance to wheelchair users at home.

\begin{figure}
    \centering
    \includegraphics[width=0.6\linewidth]{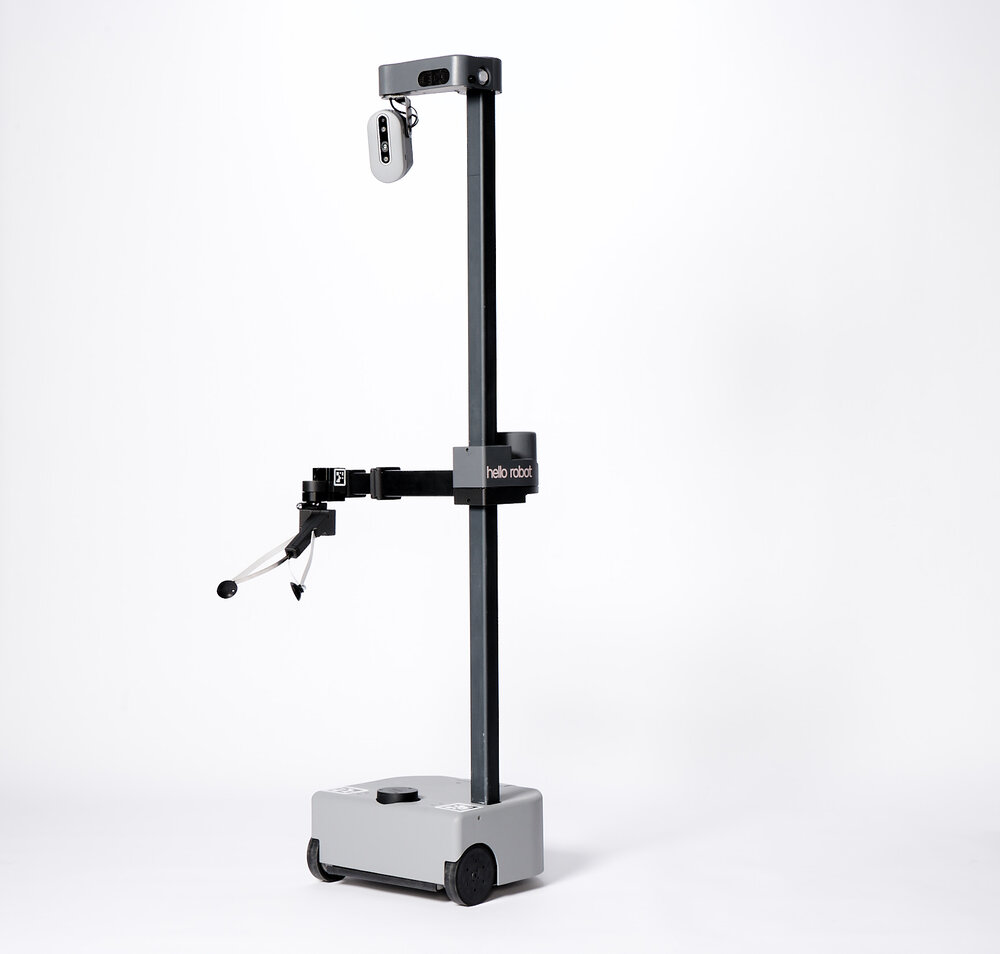}
    \caption{Stretch RE1 robot.}
    \label{fig:strecth}
\end{figure}

This paper aims to develop a pipeline that can help wheelchair
users with their daily life tasks. We implemented the WeHelp system on 
Stretch Research Edition robot (RE1) by Hello Robot \footnote{Hello
Robot, 2020; \url{https://hello-robot.com/product}} - which is a novel mobile manipulator designed for domestic settings, as shown in \cref{fig:strecth}. Our WeHelp system is composed of
five modules: speech recognition, visual tracking, mode switch-
ing, remote control interface, and teleoperation interface. The speech recognition module is used to recognize speech commands from wheelchair users. The visual tracking module served as a wheelchair follower on the behind or alongside the wheelchair user. Mode switching is used to change the following modes (following from behind or following in accompany mode). In the remote-control function, a caregiver will take over the robot through the remote-control interface. In the teleoperation function, the wheelchair user takes over the robot themselves to finish their desired tasks. 

\section{Related Work}
\label{sec:related_work}

\subsection{Shared autonomy system for wheelchair users}
\begin{figure}[htp]
    \centering
    \includegraphics[width=0.4\linewidth]{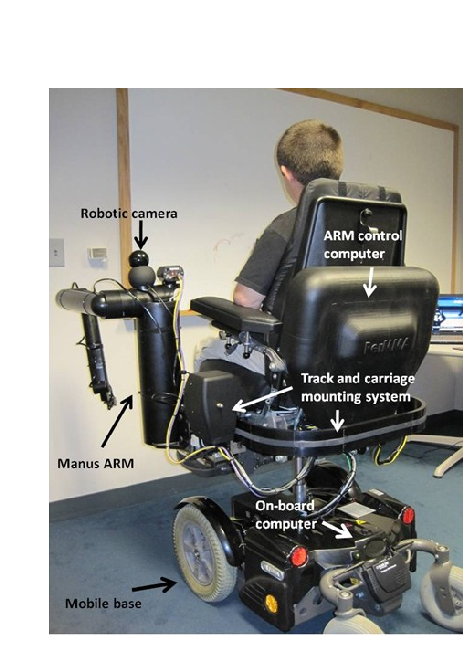}
    \caption{PerMMA (Gen I).}
    \label{fig:PerMMA}
\end{figure}

Although not many studies developed a shared autonomy system for wheelchair users, many assistive robots are integrated with wheelchairs, giving insights into our system design. One example of the personal assistive robots— PerMMA (Gen I) shares a similar design as our system \cite{wang2012personal}, shown in \cref{fig:PerMMA}. Various sensors are integrated to give users real-time feedback. The web camera on the arm provides visual feedback to the remote operator so that the visual information perceived by the user can be synchronized with the caregiver. Our study would also apply computer vision methods to human tracking. The robot can capture the movement of the user. Moreover, PerMMA (Gen I) had two user interfaces touchpad and remote control station, and three control modes-local user, remote user, and cooperative control. This feature of designs is useful to our caregiver system as well. If the caregiver is in the same place as the user, it would be easier to control the wheelchair by directly using the touchpad. The remote control is more beneficial when they are in separate places. The cooperative control mode is well-designed for a variety of scenarios. For instance, the patient himself might have some mobility in which he can use the touchpad independently. When difficult tasks are encountered, he can simply switch to control remotely by the caregiver. 

\begin{figure}
    \centering
    \includegraphics[width=0.4\linewidth]{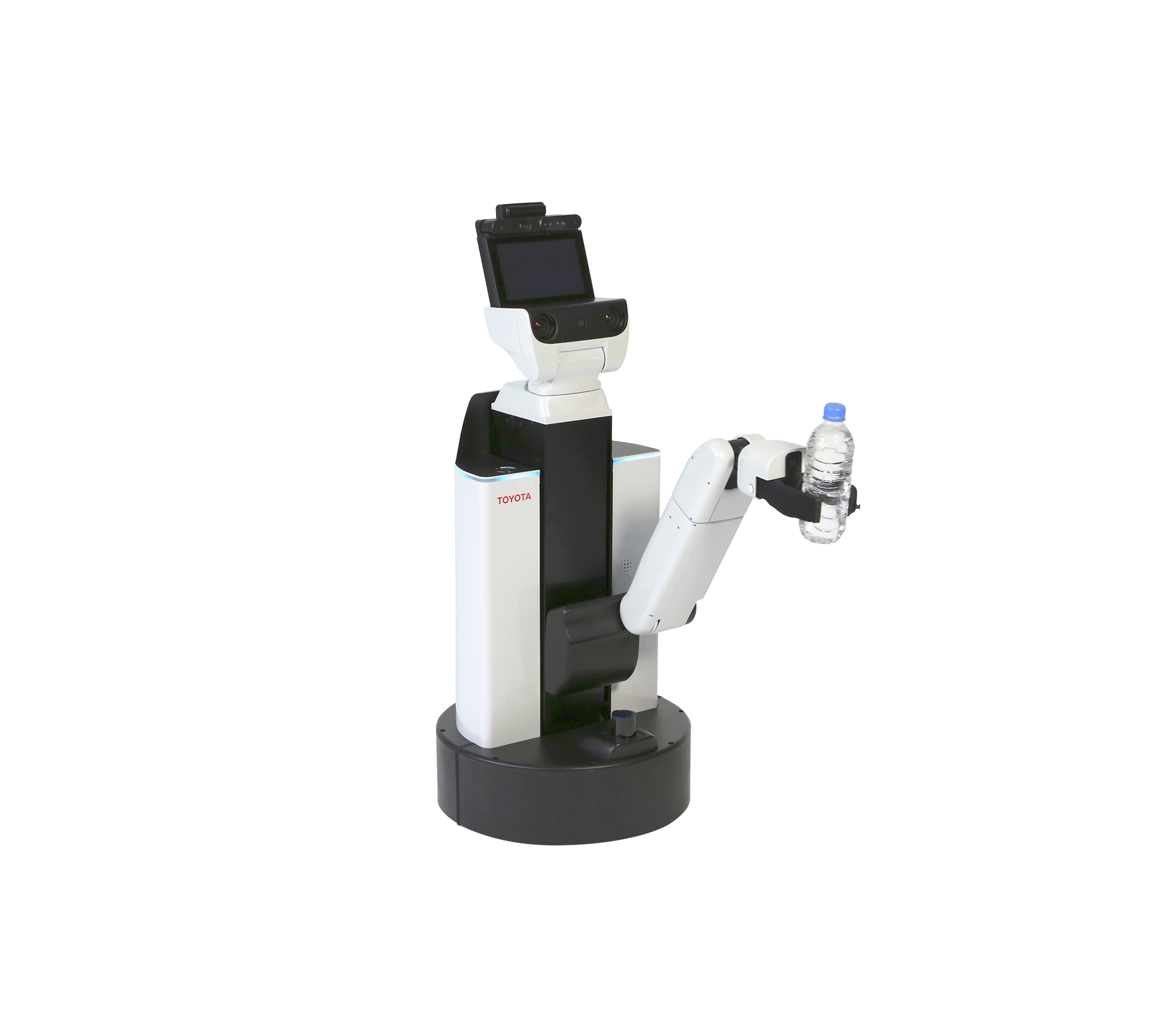}
    \caption{Toyota Human Support Robot.}
    \label{fig:HSR_robot}
\end{figure}

Another robot developed by Toyota-- Human Support Robot (HSR) can be operated by voice command or by tablet PC aligned with the purpose of our robotic caregiver system \cite{yamamoto2018human}. HSR has a lightweight body with a wheeled base (over 4 feet tall), a folding arm with a gripper, and a touchscreen controller, shown in \cref{fig:HSR_robot}. These features enable it to pick up objects off the floor, retrieve objects from high locations, open curtains, and perform other household tasks. HSR is also able to fetch and pick up thin objects using its multi-DOF arm while avoiding obstacles. All the features of HSR are very similar to our design, which could be the achievable goal of our shared autonomy system. HSR was successfully able to fetch water and open doors for a quadriplegic first tested in the United States in 2017. However, if it is considered to be launched commercially, the cost is still a big issue, which is currently priced at about 9,000 dollars on top of the 3,000 dollars monthly fee. 

\subsection{Visual Tracking}
In our work, we use a visual tracking module to follow the wheelchair user. Visual tracking is an important problem in computer vision which has a wide range of applications. Modern visual tracking systems usually follow the tracking-by-detection paradigm \cite{wang2020towards}. It has 1) a detection model for object (or human) detection from images and 2)  connecting detected objects from the video stream. Among them, efficient object detection is a challenging problem.  

A modern object detection model is usually composed of two parts, a backbone which is pre-trained on ImageNet, and a head which is used to predict classes and bounding boxes of objects. The backbone of the model could be VGG \cite{simonyan2014very}, ResNet \cite{he2016deep} , or MobileNet \cite{howard2017mobilenets}. As to the head part, it is usually categorized into two kinds, i.e., one-stage object detector and two-stage object detector. The most representative two-stage object detector is the R-CNN \cite{girshick2014rich} series.  As for the one-stage object detector, the most
representative models are the YOLO series \cite{redmon2016you,redmon2017yolo9000,redmon2018yolov3,bochkovskiy2020yolov4}. Among them, YOLO-v3 \cite{redmon2018yolov3} is a fast and accurate real-time object detection framework. Thus we used YOLO-v3 \cite{redmon2018yolov3} based approach on our visual tracking module. In several studies, visual tracking of human motions has been utilized to predict human intentions and facilitate human-robot collaboration by analyzing the tracked trajectory \cite{abuduweili2021robust,abuduweili2023online}. Building on this approach, our work employs human trajectory tracking to determine the optimal positioning of assistive robots operating in accompanying mode.

\subsection{Speech Recognition}

We used an end-to-end speech recognition module to recognize acoustic keywords, then used these keywords to activate different functions of our WeHelper system. Over the past decades, a tremendous amount of research has been done on the use of deep learning for speech recognition \cite{nassif2019speech}. 
The conventional speech recognition systems are based on representing speech signals using Gaussian Mixture Models (GMMs) that are based on hidden Markov models (HMMs). However, they are considered statistically inefficient for modeling non-linear or near non-linear functions. Opposite to HMMs, deep learning-based methods permit discriminative training in a much more efficient manner \cite{halageri2015speech}. Deep Speech is an efficient and accurate end-to-end speech recognition system \cite{hannun2014deep}. The architecture of Deep Speech is significantly simpler than traditional speech systems. Because Deep Speech replaces entire pipelines of hand-engineered components of traditional methods with neural networks, end-to-end learning allows us to handle a diverse variety of speech including noisy environments, accents, and different languages. Deep Speech 2 is an upgraded version of the original Deep Speech system with more accurate performance \cite{amodei2016deep}.  Thus we used Deep Speech 2 \cite{amodei2016deep} based approach on our speech recognition module.

\section{Stakeholders}
We have talked to our stakeholders, a 70-year-old wheelchair user, and her daughter (caregiver). She had a stroke last year which impairs her mobility to move the arms and legs. Thus, she uses the wheelchair on a daily basis. During the design of our project, we conducted an interview, especially learning about her special needs as a wheelchair user and the role of a caregiver in her situation. First, she stated that assistance in small things is essential including getting the water bottle, going to the bathroom, taking a shower, etc. Furthermore, she supported this by illustrating the benefits of using the robotic caregiver. For example, going to the restroom is a private activity, and usually, she said, “it’s kind of embarrassing to go to the bathroom with someone unknown (human caregiver) besides family members.” By contrast, she thought, “Having a robotic caregiver will make me feel more comfortable and less embarrassed”. 

Second, although caregivers are needed in many scenarios, independence is a wheelchair user’s top priority. Her daughter said, “Once the caregiver like me is not at home, she needs to do things on their own.” Thus, the user needs to control the robot caregiver to do something more complex if needed. Give an example of opening the door, it is a difficult task for the users especially since some doors are open toward them requiring some space between the wheelchair and the door. This task is hard to preprogram since doors are different, and so in this situation, the caregiver is better off controlling the robot manually and opening the door.

\section{Method}

\begin{figure*}
    \centering
    \includegraphics[width=0.7\linewidth]{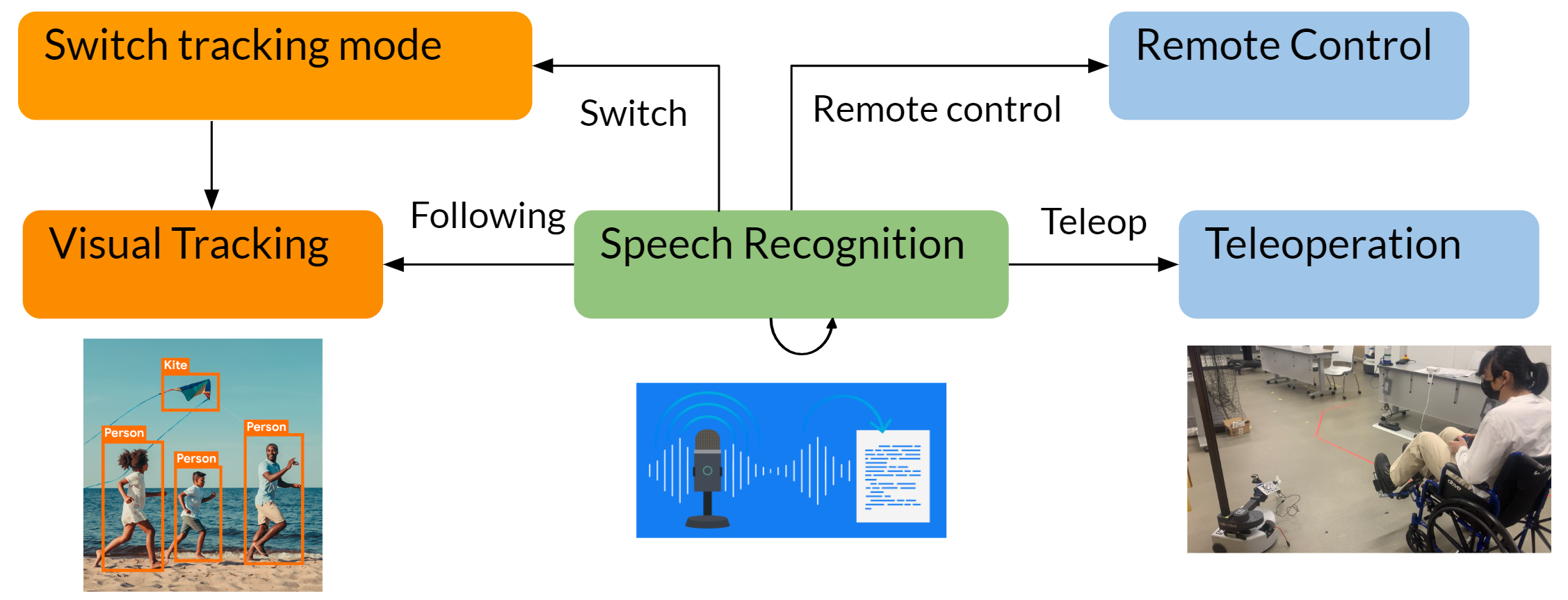}
    \caption{Illustration of the method pipeline. The speech recognition module decides which module is active.}
    \label{fig:pipeline}
\end{figure*}

We aim to develop a pipeline that can help the wheelchair user with their daily life tasks. Our method is composed of five parts: speech recognition, visual tracking, mode switching, remote control interface, and teleoperation interface. The robot has three modes: following mode, remote control mode, and teleoperation mode. The speech recognition module decides which mode is going to be activated. When the following mode is activated, the robot follows the wheelchair user automatically through a computer vision-based tracking algorithm. Besides, the user can ask the robot to follow them from different angles, including from behind, by the left, and by the right. When the remote control mode is activated, a caregiver will take over the robot through the remote control interface and assist the wheelchair user in completing complex tasks. When the teleoperation mode is activated, the wheelchair user takes over the robot themselves and controls the robot to finish their desired tasks. The pipeline is shown as \cref{fig:pipeline}.

In the next section, we will introduce how we implement these five modules.

\subsection{Speech recognition}
The speech recognition module is composed of two major parts: 1) iStreach robot microphone setup for environmental sound monitoring; 2) the real-time speech sentence detection; 3) microphone speech-to-text translation; and 4) remote control triggering system based on key words detection. Following we will introduce the four parts in sequence. 

First, the most important step is to access the microphone module of the iStream robot, such that the speech recognition module can monitor the environmental sound in real-time. The major tool we use is called Pyaudio, which provides the Python bindings for PortAudio, the cross-platform audio I/O library. Therefore, we can use Python to play and record audio on the iStreach platform (Ubuntu/Linux). Specifically, We leverage the Pyaudio library to identify the accessible devices in the iStreach robot, where we directly use the default sound card as well as the default microphone as the target device. Once the target machine address is assigned, we then create a Pyaudio streaming object, which keeps logging the environmental sound stream frames that go through the microphone.

Second, once we can access the microphone I/O, we will control the microphone-detected speech information to smartly identify the possible sentence from the user. Note that the microphone is constantly working to monitor the environmental sounds, whereas it is desired to automatically detect when the user's speech is beginning and when the uttering is ended, such that we can pass the complete sentence for voice recognition. Specifically, we leverage the toolbox called WebRTC Voice Activity Detector (VAD), which automatically classifies a piece of audio data as being voiced or unvoiced. Therefore, we constantly use the audio to detect audio frames, once the voiced frame from the record hits a certain length, we will send the concatenated frames for voice recognition while ignoring the rest of the received frames during speech recognition. We will only restart voiced frame buffering once the unvoiced frames hit a certain length, i.e. the user stops talking, and hence a new sentence recording can be started.

Third, once we have the voiced speech frames, we then translate the audio information into the text information, such that the key words detection module can be triggered to filter the keywords. Our key audio-to-text translation model is based on the speech recognition system called deep Speech from Baidu Research~\cite{amodei2016deep}. The deep speech model is an end-to-end speech system, where deep learning supersedes these processing stages. Specifically, the Deep Speech system uses recurrent neural networks to map well to GPUs and a model partition scheme is used to improve parallelization. Overall, the advantage of the capacity provided by deep learning systems is leveraged to learn from large datasets to improve the overall speech recognition performance. With both collected and synthesized data used during training, Deep Speech exhibits robust good performance in the presence of realistic noise and speaker variation.

Finally, with the translated text from the user's speech, we construct a keyword detection system to trigger the remote control when necessary. With well-detected words, the keyword detection system is relatively easy to construct. Specifically, the system monitors whether a certain combination of keywords exists in the detected text sequence; if the keyword is detected, then the trigger system will start the visual tracking process as well as the real-time audio text translation process, and the human assistant will take control of the robot. Keywords and related commands are shown in \cref{tab:keywords}.

Note that the default microphone of iStreach robot is located in the bottom wheel of the robot, which is highly unreliable since the robot is constantly moving and a large background noise exists to interfere with the recognition accuracy. To address this problem, we connect an external microphone to the proposed system and tune the sound card input sampling rate into a higher frequency, i.e. $48000 HZ$. Through this approach, we can monitor the high-quality sound from the user by installing the external microphone on top of the side of the wheelchair.

\begin{table}[h]
\begin{tabular}{l|l}
\hline
Keywords       & Commands and Results                                                       \\ \hline
go left        & follow the wheelchair user on left                                        \\
go right       & follow the wheelchair user on right                                       \\
go back        & follow the wheelchair user from behind                                    \\
remote control & teleoperation mode, wheelchair user can control the robot \\
help           & remote assistance mode, caregiver can control the robot   \\ \hline
\end{tabular}
\caption{Keywords used in speech recognition module.}
\label{tab:keywords}
\end{table}

\begin{figure}
    \centering
    \includegraphics[width=0.9\linewidth]{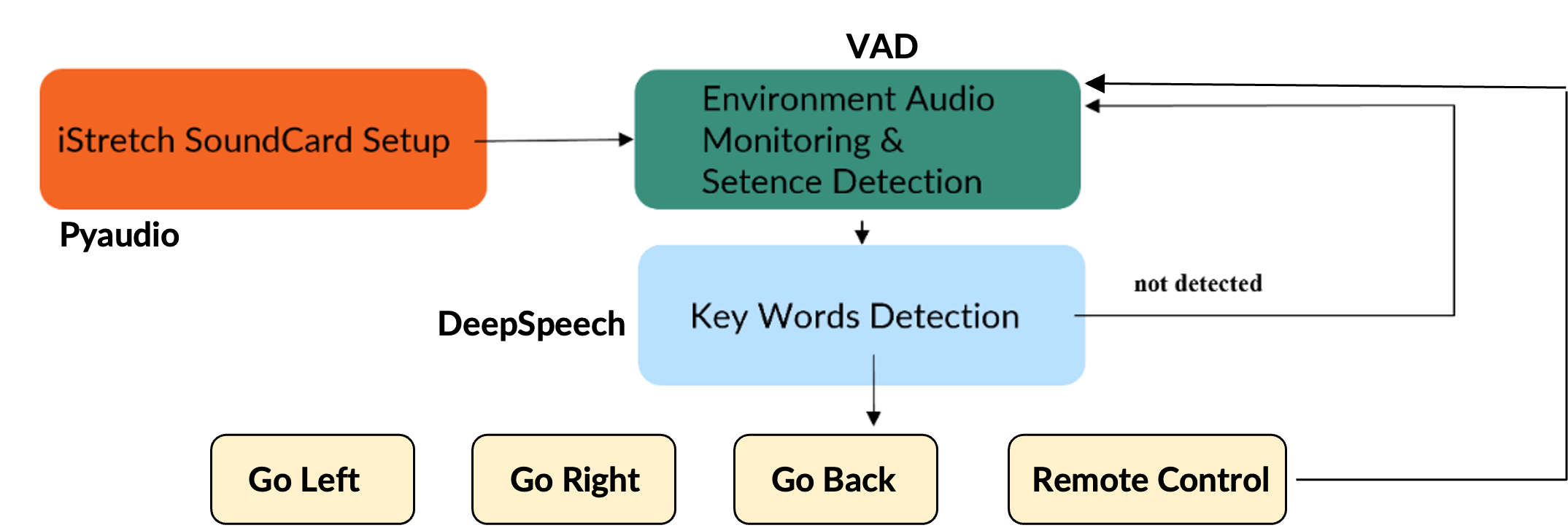}
    \caption{Illustration of speech recognition module.}
    \label{fig:speech}
\end{figure}

To better illustrate how the overall speech recognition module works, we provide an overview illustration in  \cref{fig:speech}.

\subsection{Visual tracking}

\begin{figure}
    \centering
    \includegraphics[width=\linewidth]{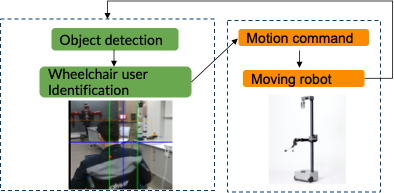}
    \caption{Visual tracking architecture. }
    \label{fig:track}
\end{figure}

The visual tracking module includes four parts: object detection, wheelchair user identification/filtering from detected objects, motion command generation to follow the human, and moving robot. Each part of the visual tracking module is shown in \cref{fig:track}. 

\subsubsection{Object detection}

\begin{figure}
    \centering
    \includegraphics[width=\linewidth]{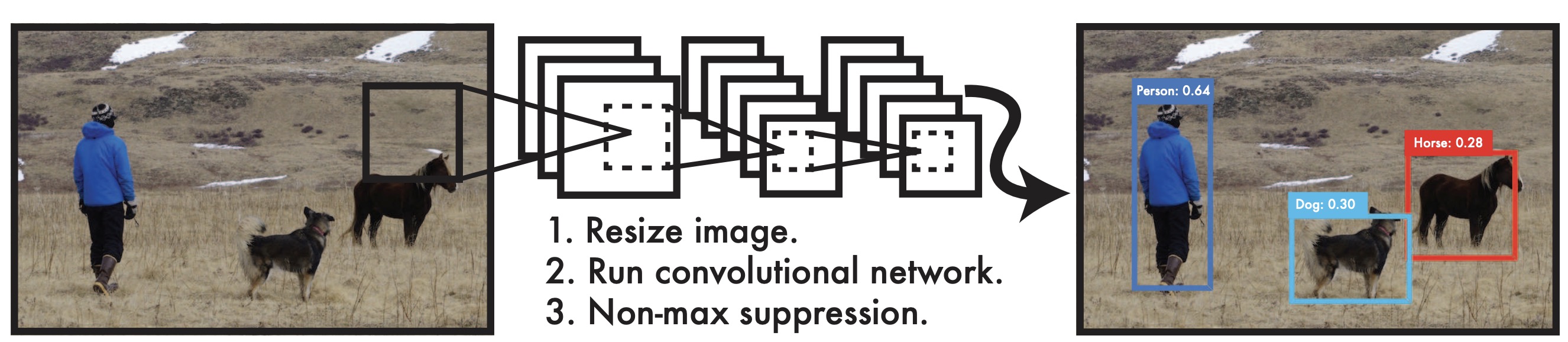}
    \caption{The YOLO Detection System. Processing images
with YOLO is simple and straightforward. }
    \label{fig:yolo}
\end{figure}

Our object detection model is based on YOLO \cite{redmon2018yolov3}. YOLO (You Only Look Once) is a real-time object detection framework \cite{redmon2016you}. Which predicts bounding boxes of objects and class probabilities directly from full images in one evaluation, as shown in \cref{fig:yolo}. The most important advantage of YOLO is its fast processing speed. Since YOLO frame object detection is a regression problem, we don’t need a complex pipeline. We simply run our YOLO neural network on a new image at inference time to predict detections.  
In our implementation, we used the YOLO-v3-tiny model \cite{redmon2018yolov3}. YOLO-v3 is more accurate than the original YOLO. And we did not train the YOLO-v3-tiny model by ourselves, we directly used the pretrained model on COCO datasets \cite{lin2014microsoft}. 
In our implementation, the object detection model achieves $2 \sim 3$ fps to process camera video on the Stretch RE1 robot.

\subsubsection{Wheelchair user identification}

As described in the above section, we used the COCO-pretrained YOLO-v3 model as our object detector. COCO-pretrained YOLO-v3 model could detect 80 common objects but could not detect wheelchairs, and there are no open-source object detection models to detect wheelchairs. Thus, we use the following trick to identify Wheelchair users from all detected objects. Firstly, we used the COCO-pretrained YOLO-v3 model to detect all the humans in the camera. Secondly, we select the human with the larger bounding box and the shorter height. Because wheelchair users are always close to the camera, and wheelchair users are always shorter than walking people in the background. We acknowledge that this trick is not very robust, but accurate and simple enough for many applications. The best approach to detecting wheelchair users is to train from scratch an object detection model with wheelchair data. But our resources don't allow it for now, so we didn't train the model with wheelchair data.

\subsubsection{Motion command generation}

\begin{figure}
    \centering
     \begin{subfigure}[b]{0.4\textwidth}
     \centering
     \includegraphics[width=0.7\textwidth]{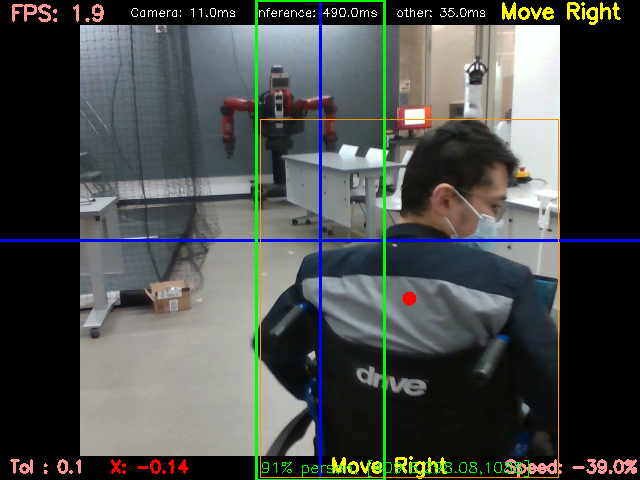}
     \caption{Following from behind. Command: move right}
     \label{fig:move_right}
 \end{subfigure}
     \begin{subfigure}[b]{0.4\textwidth}
     \centering
     \includegraphics[width=0.7\textwidth]{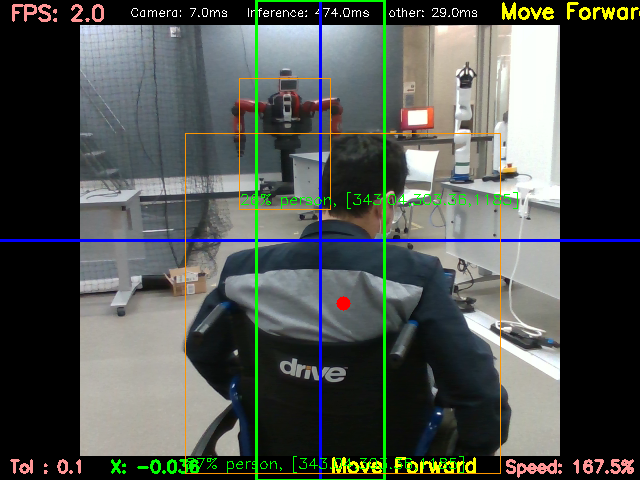}
     \caption{Following from behind. Command: move forward}
     \label{fig:move_forward}
 \end{subfigure}
    \caption{Robot's view during visual tracking process (Following from behind). }
    \label{fig:track_move}
\end{figure}

\begin{figure}
    \centering
        \begin{subfigure}[b]{0.4\textwidth}
     \centering
     \includegraphics[width=0.7\textwidth]{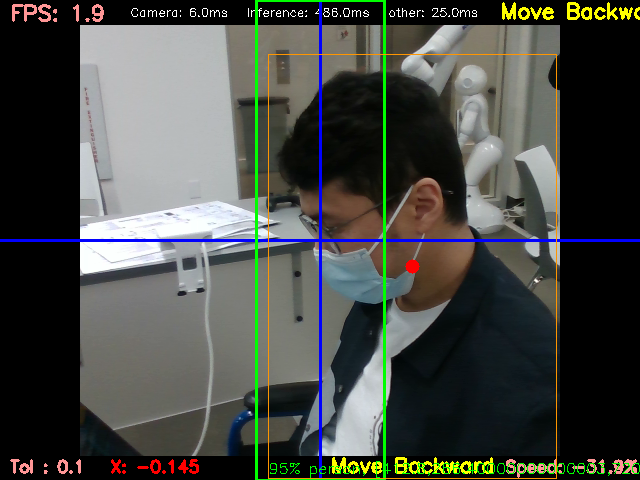}
     \caption{Following on  left. Command: move backward}
     \label{fig:left_backward}
 \end{subfigure}
     \begin{subfigure}[b]{0.4\textwidth}
     \centering
     \includegraphics[width=0.7\textwidth]{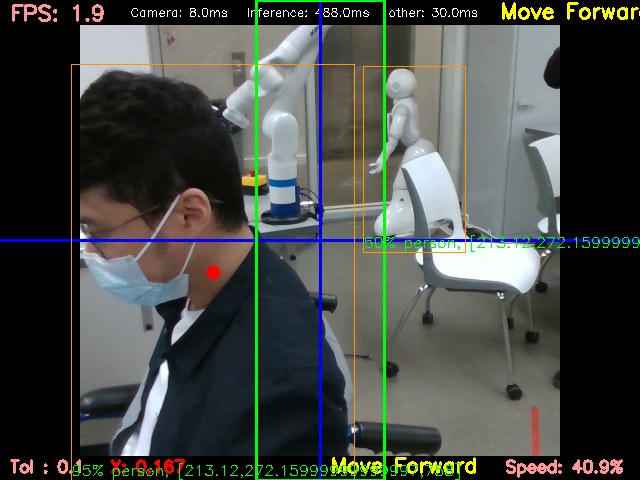}
     \caption{Following on  left. Command: move forward}
     \label{fig:left_forward}
 \end{subfigure}
    \caption{Robot's view during visual tracking process (Following on  left). }
    \label{fig:track_left}
\end{figure}

The Robot follows the wheelchair user and maneuvers itself to get the wheelchair user in the center of the frame. We describe the approach to determining transition and rotation values of robot motion as below.
\begin{itemize}
    \item Following from behind the wheelchair user. In this mode, a robot follows the wheelchair user from behind. We used a depth camera to calculate the distance between the robot and the wheelchair user. Then the distance to determine the transition (forward and backward) of the robot. The rotation of the robot is determined by the deviation of the center of the wheelchair user from the center of the camera frame. While the robot is tracking/following the human, the examples of the Robot's view are shown in \cref{fig:track_move}. The robot's view with information overlay is generated using OpenCV. When the human is present in the frame, information such as bounding boxes (orange curved rectangular), the center of the human (red dot), deviation of the object from the center of the frame (blue cross), robot direction, and speed are updated as shown in picture \cref{fig:track_move}. In the example \cref{fig:move_right}, the X value denotes the deviation of the center of the wheelchair user (the red dot) from the center of the frame. Since the horizontal deviation (i.e. value of 'X') is above the tolerance value (that means the wheelchair user is too right), the code generated 'Move Right' command. In the example \cref{fig:move_forward}, the distance between the robot and the wheelchair user is larger than the tolerance value, and the code generated 'Move forward' command.
    
    \item Following with accompanying mode. In this mode, a robot follows the wheelchair user by the right or left. In the accompanying mode, there is an angle of 90/-90 degrees between the face angle (camera direction) and base angle (base motion direction) of the robot.  Different from the following-from-behind mode, in the accompanying mode, the transition of the robot is determined by the deviation of the center of the wheelchair user from the center of the camera frame. Take the following on left mode as an example. As shown in \cref{fig:left_backward}, horizontal deviation (i.e. value of 'X')  is above the tolerance value (that means the wheelchair user is on back), and the code generated 'Move Backward' command.
\end{itemize}

\subsubsection{Moving robots}

After generating motion commands, we directly use Python script to push the moving command to the robot. As long as the speech recognition is triggered by keywords (e.g. "remote"), the visual tracking module stops immediately, otherwise, the tracking continues.

\subsection{Mode switching}

We allow the user to ask the robot to follow them from different angles, including following from behind, following by the left, and following by the right. This module deals with the switch between different following modes.

\begin{figure}
    \centering
    \includegraphics[width=0.8\linewidth]{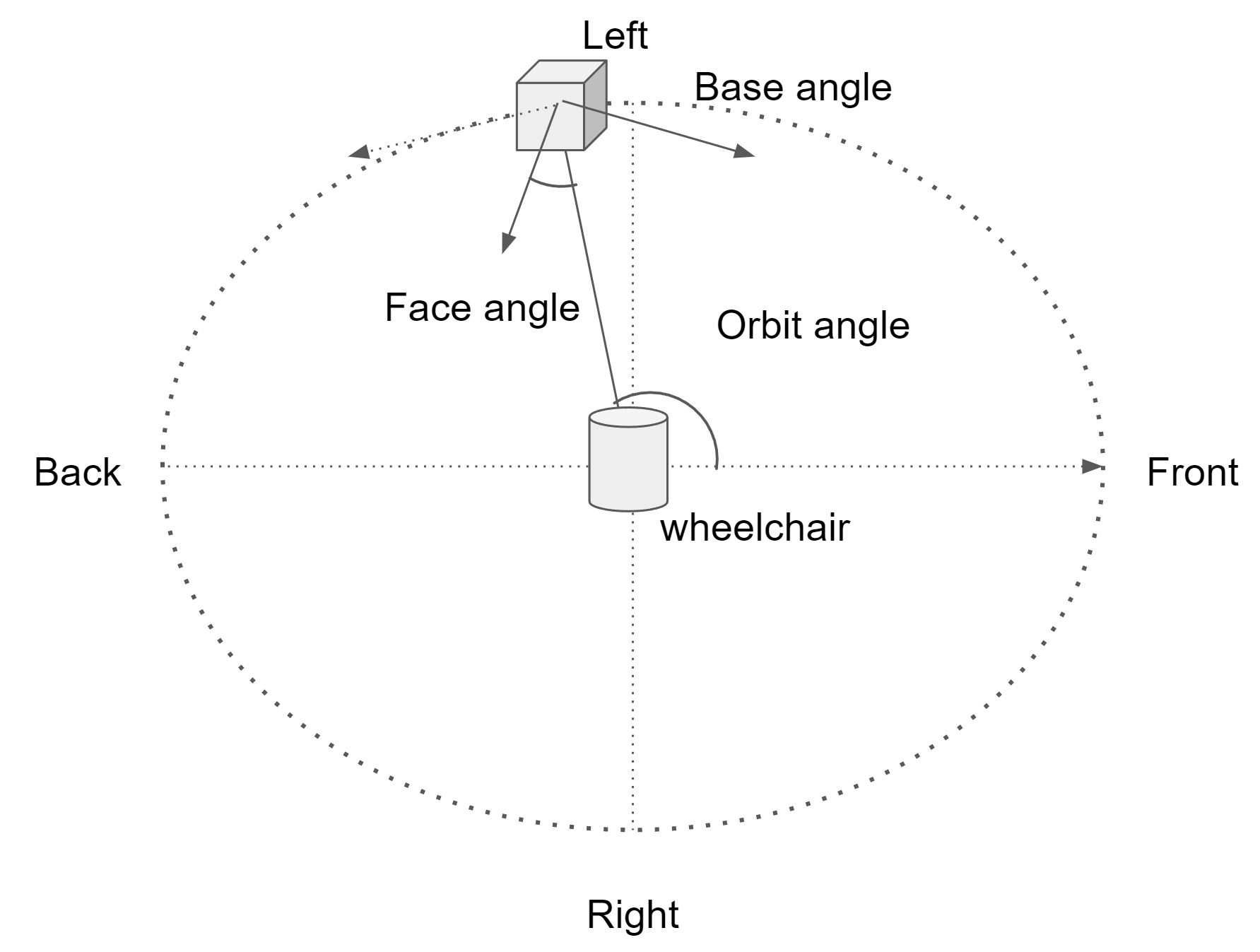}
    \caption{Illustration of mode switching concepts.}
    \label{fig:switch}
\end{figure}

To enable the robot to switch between different modes, we need to adjust the distance between the robot and the wheelchair user, the angle of the robot base (base angle) and the angle of the camera (face angle) as shown in \cref{fig:switch}. And we also need to generate the switch trajectory.

Before the switch, we need to know the orbit angle of the robot, the base angle of the robot, and the distance from the robot to the human. If we have a QR code sticker on each side of the wheelchair, this information can be easily inferred from the camera transformation. We implemented a way that does not rely on the QR code. We estimate the distance from the robot to the human directly from the depth camera. We estimate the base angle and orbit angle from historical motion data. Except for these measures, to generate the switch trajectory, we also need to define the target base angle, target face angle, and target distance between the robot and the wheelchair. We define these in advance for all following modes as shown in \cref{tab:target_pose}.
\begin{table*}[h]
    \centering
    \begin{tabular}{c | c c c c}
    \toprule
         Mode & Target Distance (m) & Target Orbit angle & Target Base angle & Target Face angle \\
         \midrule
         Following by left & 0.5 & 90 &  0  &  270\\
         Following from behind & 1.2 & 180 & 0 &  0\\
         Following by right & 0.5 & 270 &  0  &  90\\
         \bottomrule
    \end{tabular}
    \caption{Mode switching parameters}
    \label{tab:target_pose}
\end{table*}

Next, we define the switch trajectory from these measures. Here we only consider two cases ``left to behind" and ``right to behind". Because other cases are either the reverse process of these two cases or a combination of these two cases and their reverse process. For example, ``left to right" can be decomposed as ``left to behind" and ``behind to right". We first define two midpoints for ``left to behind" and ``right to behind" separately. Then we generate a two-step trajectory composed of two straight-line move connecting with the midpoint for each case. The reason we do not use a one-straight trajectory is that it may lead to a collision between the robot and the wheelchair.

\subsection{Remote control}

\begin{figure}
    \centering
    \includegraphics[width=0.9\linewidth]{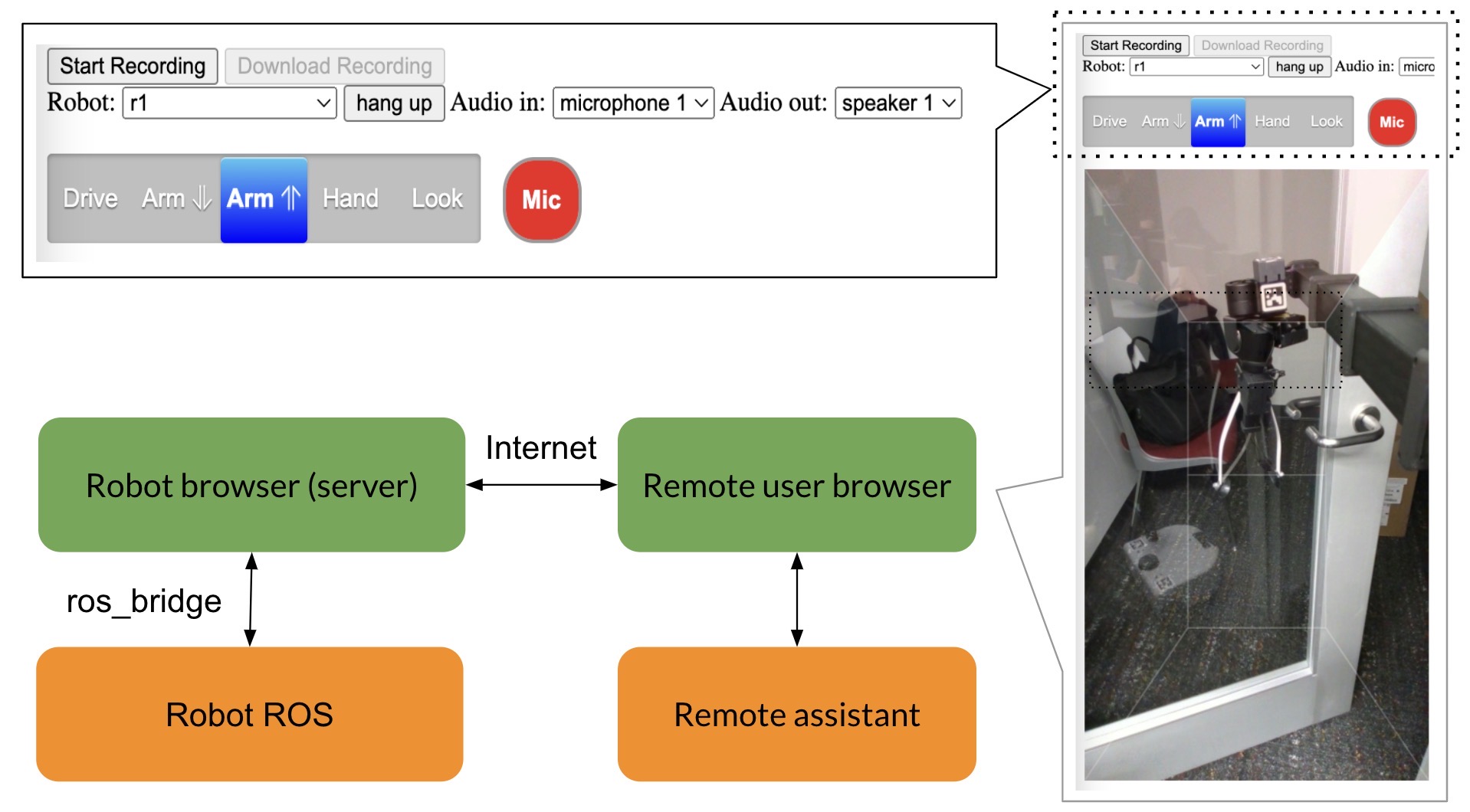}
    \caption{Remote control architecture. The remote control is based on WebRTC (Web Real-Time Communication). The robot and the remote operator can communicate via browsers on the Internet. The remote operator can receive video and audio streaming. }
    \label{fig:remote}
\end{figure}

The remote control is based on the Stretch Web Interface. A web-based package for remote access of the stretch robot. The package is developed based on WebRTC (Web Real-Time Communication). A well-tested free open-source platform that provides web browsers and mobile applications with real-time communication via APIs. A server is established in a browser on the robot, then remote operators can access the server via the Internet on their devices. The operator can receive real-time video and audio streaming and send desired operation commands. The interface is shown in \cref{fig:remote}.

To start the teleoperation, the first step is to establish a server on the robot. If it is a production environment, we could use an external server instead of the robot to handle connections and teleoperations. The server uses the Express web framework and provides a WebRTC signaling service using sockeit.io. The sessions are stored by Redis. The identity of the remote operator is verified using Passport.js, Mongoose, and MongoDB. Once connected, the operation interface pops out. 

The operation interface has 5 tabs as shown in \cref{fig:remote}. In each tab, the operator can control the robot by clicking different areas of the image. In the first tab, the operator can control the mobile base to move the robot. In the second tab and the third tab, the operator can move the robot arm up and down and the orientation of the arm. But the camera faces a lower angle in the second tab and a higher angle in the third tab. The operator can control the manipulator in the fourth tab and rotate the camera in the fifth tab.

The command is packed as JSON data and sent to the robot browser. Then the robot browser uses rosbridge to connect with ROS on the robot. Rosbridge translates JSON from the robot’s browser into ROS communications and vice versa.

\subsection{Teleoperation}

The teleoperation module is based on Stretch Body. This open-sourced package provides an interface to teleoperate the stretch RE1 robot via Xbox controllers. We programmed an interface to connect Stretch Body teleoperation tools with our pipeline. When the speech recognition module receives a command to switch to the teleoperation mode, we will instantiate an Xbox controller instance and enter teleoperation mode. We mapped the ``X" button on the controller to exit teleoperation mode. So when the wheelchair user finishes their tasks with the robot, they can switch the robot back to the following mode by simply pressing the ``X" button.

\section{Evaluation}
\label{sec:evaluation}
We evaluate our WeHelp system in two manners: 1) separately evaluate the speech recognition and visual tracking parts. 2) evaluate the operation time and success rate for daily life tasks with shared autonomy. 

\subsection{Evaluation of speech recognition module}
To evaluate the performance of the speech recognition module, we designed a set of experiments to evaluate the robustness of speech recognition under different environmental noise levels. Note the most important feature of speech recognition is to ensure that accurate recognition can maintain a high success rate even when the environmental noise is very large, due to the fact that the patient may operate the proposed wheelchair helper system in a public area, where noise is not negligible at all. 

To mimic levels of background noise level, we design two sets of experiments with different background sound decibel levels, including $50 db, 60db$. The background noise is controlled by controlling the distance between a sound source and the microphone. The sound source uses the youtube sound clip from the thunderstorm. We make sure the decibel value at the microphone is accurate through testing with the iPhone decibel detection module. To ensure the robustness of the recognition success rate across the users individual differences, we invite four human subjects to 
repeat the speech recognition experiments. There are four keywords included, i.e. "go to the left", "go to the right", "go back", and "remote control". For each human subject, he/she will repeat each keyword 20 times for one environmental sound noise level. Therefore, in total, we have $20 \times 4 \times 2 = 160$ test samples. 

We summarize the detailed speech recognition success rate for each human subject under different background noise levels in \cref{tab:speech results}. The comparison results in \cref{tab:speech results} indicate that 1) speech recognition is robust to environmental noise, and 2) the recognition success rate highly depends on the accent of the accent of the patient. 

\begin{table}[h]
    \centering
    \begin{tabular}{c|c|cccc}
    \toprule
         Environmental Noise & Subject & left & right & back & remote\\
         \midrule
         & S1 &  $65\%$ & $100\%$ & $65\%$ & $60\%$ \\
    50db & S2 &  $65\%$ & $95\%$ & $80\%$ & $45\%$\\
         &S3 &  $5\%$ & $50\%$ & $95\%$ & $15\%$ \\
         & S4 &  $95\%$ & $95\%$ & $100\%$ & $30\%$ \\
         \midrule
         & S1 &  $70\%$ & $95\%$ & $100\%$ & $60\%$ \\
    60db & S2 &  $45\%$ & $80\%$ & $85\%$ & $30\%$\\
         &S3 &  $0\%$ & $50\%$ & $55\%$ & $6\%$ \\
         & S4 &  $85\%$ & $60\%$ & $100\%$ & $15\%$ \\
    \bottomrule
    \end{tabular}
    \caption{Evaluation results in terms of keyword recognition success rate under different environmental noise levels across different human subjects.}
    \label{tab:speech results}
\end{table}

\subsection{Evaluation of visual tracking module}

We evaluate the effectiveness of the visual tracking module by reporting the success rate of users following with different wheelchair speeds. \Cref{tab:evalfollow} shows the success rate of the human-following module under different wheelchair user speeds and different following modes (following from behind, following on the right, following on the left). We can draw some conclusions from the table. 1) The success rate of human following relies on the speed of the wheelchair user, especially for following on the right and left. The main reason is that our Stretch RE robot moves at a slow speed, making it easy to lose track of the wheelchair user. 2) Following from behind performs much better than following in the accompaniment mode because following from behind offers a better view (wider camera vision) and more flexible movement (easier and faster forward motion of the robot’s base). 3) Following on the left performs better than following on the right because when tracking from the right, part of the camera's field of view is blocked by the vertical column of the robot itself, as shown in \cref{fig:strecth}.
 
\begin{table}[h]
\centering
\begin{tabular}{l|lll}
\hline
Speed ~\textbackslash{} Mode & Behind & Right & Left  \\ \hline
0.1 m/s                   & 100\%  & 100\% & 100\% \\ \hline
0.2 m/s                   & 100\%  & 60\%  & 85\%  \\ \hline
0.3 m/s                   & 100\%  & 20\%  & 30\%  \\ \hline
1.0 m/s                   & 100\%  & 0\%   & 0\%   \\ \hline
\end{tabular}
\caption{Evaluation of visual tracking module.}
\label{tab:evalfollow}
\end{table}

\subsection{Evaluation of operation time}

We evaluated the performance of our pipeline based on manipulation complexity for the tasks. To evaluate how well the robot did overall, we recorded the total time it takes for the shared autonomy system to accomplish each task. Some daily life tasks and their evaluations are listed in \cref{tab:results}. As can be seen, for a new user, the operation time for daily life tasks is slower than expected. However with training, the fluent user could decrease the operation time for our WeHelp shared autonomy system. 

\begin{table}[h]
    \centering
    \begin{tabular}{c|cc}
    \toprule
         Task & First user & Fluent user\\
         \midrule
         Moving a chair &  $\sim$ 70s & $\sim$ 40s \\
         Opening doors &  $\sim$ 90s & $\sim$ 60s \\
         Retrieving objects &  $\sim$ 60s & $\sim$ 40s \\
    \bottomrule
    \end{tabular}
    \caption{Evaluation results for some daily tasks in terms of operation time.}
    \label{tab:results}
\end{table}

\subsection{Case study}

To assess the capability of implementing the system in the real world, we conducted two use cases: 1) moving a chair; and 2) opening a door. Here, we will show the applicability of our method by illustrating a case study of moving an obstacle (i.e. a chair). 

\begin{figure}
    \centering
    \begin{subfigure}[htbp]{0.35\textwidth}
     \centering
     \includegraphics[width=0.7\textwidth]{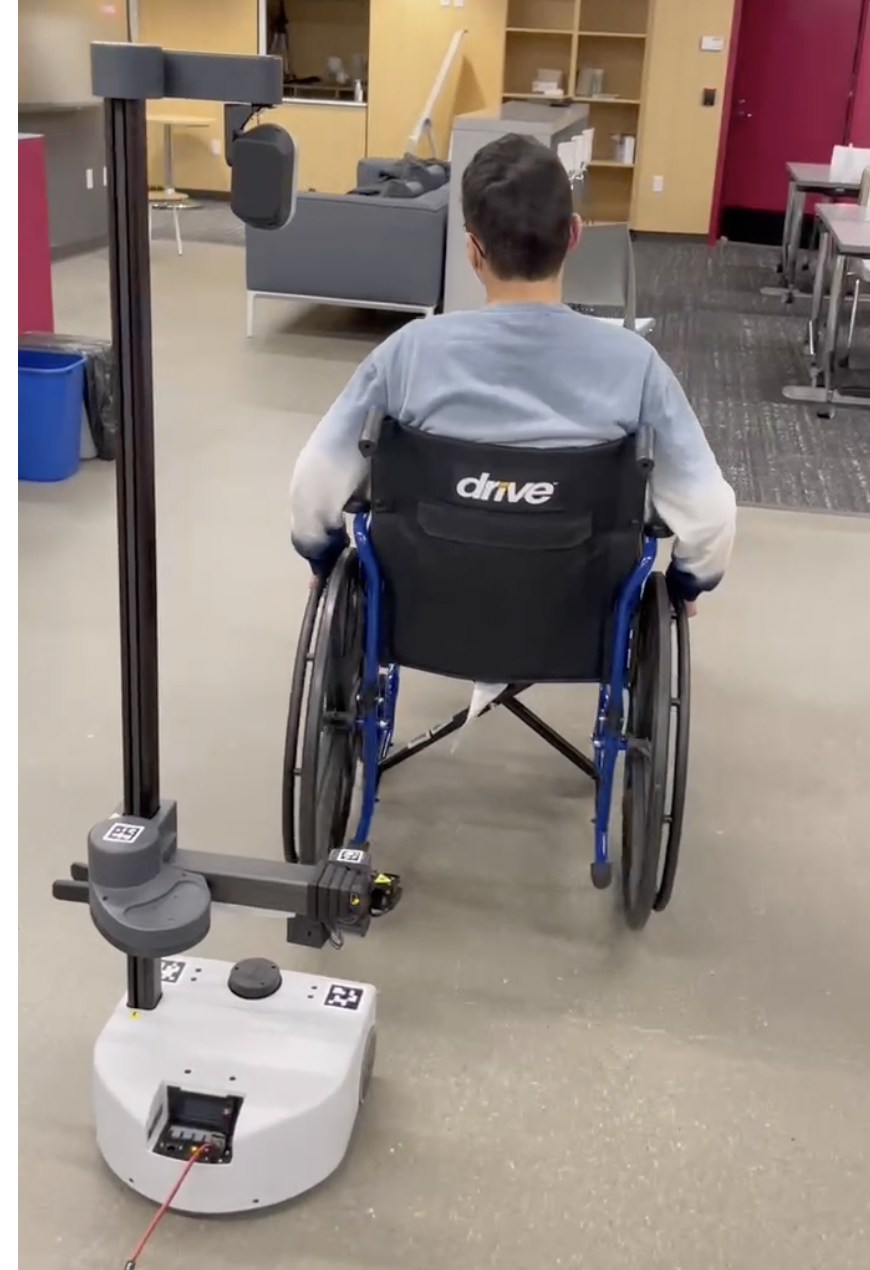}
     \caption{The following mode: the robot follows the user with a certain distance.}
     \label{fig:followingmode}
\end{subfigure}
     \begin{subfigure}[!htbp]{0.35\textwidth}
     \centering
     \includegraphics[width=0.7\textwidth]{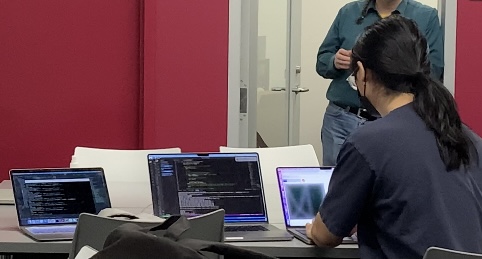}
     \caption{The remote control mode: The wheelchair user says help, is detected by speech recognition, and activates the remote control. }
     \label{fig:case-remote}
\end{subfigure}
    \begin{subfigure}[htbp]{0.35\textwidth}
    \centering
    \includegraphics[width=0.7\textwidth]{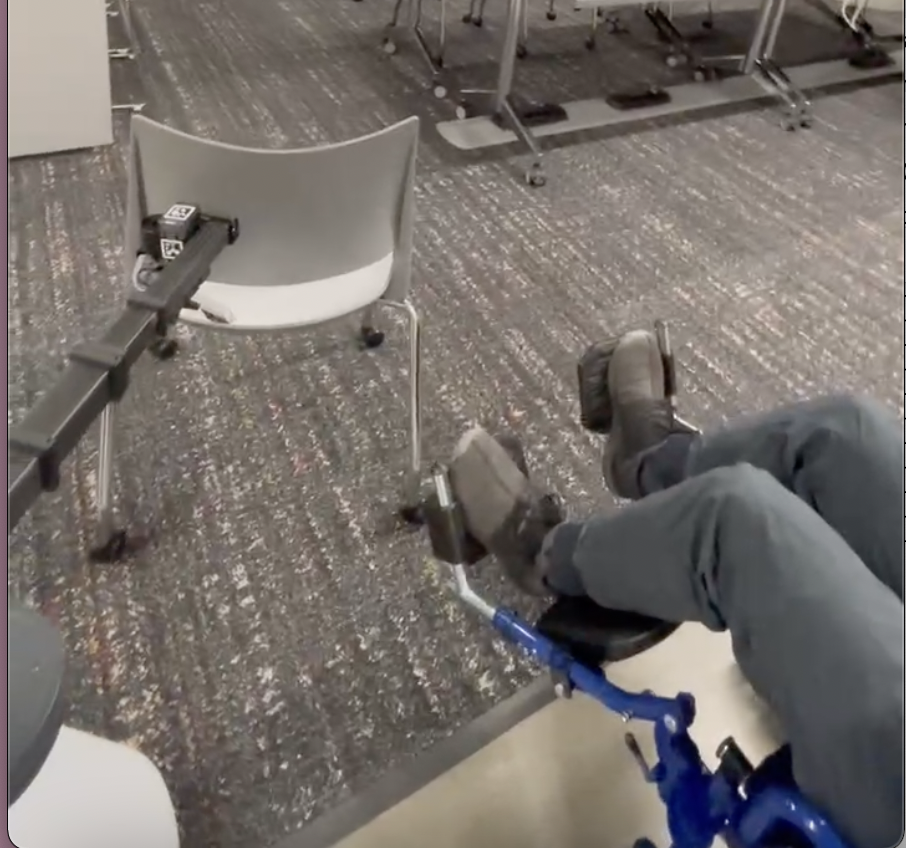}
    \caption{Moving an obstacle: The remote caregiver controls the robot and moves the chair. The wheelchair user is able to pass now. }
    \label{fig:movingchair}
\end{subfigure}
    \caption{The case study. }
    \label{fig:case_study}
\end{figure}

This case study is illustrated in \cref{fig:case_study}. The robot was initially in the following mode. By activating the visual tracking, the camera detected and recognized the wheelchair user. The robot followed the user and kept a proper distance from him. Then, an obstacle was on the way where the wheelchair user would pass. The wheelchair user said “help” to the robot. The speech recognition module recognized the command and switched the mode. The remote control mode was activated. A remote assistant took over the robot via the remote control interface and moved the chair out of the way. Finally, the task was completed, and the robot switched back to the following mode and waited for the next call.

\subsection{Feedback from stakeholdes}

We recorded some demos on some daily life tasks (i.e. moving a chair, opening a door). Our stakeholders provided feedback based on the demo: 1) it takes a long time to complete the task; 2) it is difficult to control for people with arm lifting problems. For the first problem, we need to increase the robot's speed, but it may cause safety issues. The second problem arises an important future direction, that incorporate brain-machine interface (BMI) into control.

\section{Limitations / Future Work}


The pipeline is also very brittle. The robot often gets stuck for unknown reasons. The on-board operation system Ubuntu often freezes during remote control. Because it always happens without any signal or information, we are still unclear about the actual reason. We suspect that it is caused by memory leaking or insufficient communication bandwidth. In the next, we will look into this issue.

One important future direction is exploring more effective human-robot interactions. Our WeHelper pipeline is compatible with other human-robot interactions. It is possible to use a gesture recognition module even a brain-machine interface (BMI), instead of a speech recognition module. By replacing the speech recognition With BMI, and keeping other parts, some wheelchair users with severe disability could more efficiently conduct simple daily life tasks and inquire about help from remote caregivers on complex tasks more conveniently. As discussed in feedback from stakeholders, it is also important to incorporate BMI into control. 

Another future direction is to combine visual tracking modules with trajectory prediction, collision avoidance, and SLAM. Which will make the WeHelp system more intelligent and safe.

\section{Conclusion}

In this work, we propose WeHelper -  a shared autonomy system for wheelchair users, to assist wheelchair users in daily life tasks. Our WeHelper system could follow the wheelchair user with accompanying mode and could recognize some speech commands to let the robot controller by the user or remote caregivers. The evaluation results show that the pipeline is useful for wheelchair users.  However, a mature product requires that the pipeline be smooth and robust. A significant amount of effort is required to apply the pipeline to real-world scenarios.

\bibliographystyle{unsrt}
\bibliography{bibliography.bib}

\begin{thebibliography}{10}

\bibitem{market2021north}
North america wheelchair market size, share \& trends analysis report by
  application (hospitals, homecare, ascs), by product (manual, electric), by
  category type, by country, and segment forecasts.
\newblock Technical report, Market Data Forecast, 7 2021.

\bibitem{brault2012americans}
Matthew~W Brault.
\newblock {\em Americans with disabilities, 2010: Household economic studies}.
\newblock US Department of Commerce, Economics and Statistics Administration,
  US~…, 2012.

\bibitem{wang2019artificial}
Weiyu Wang and Keng Siau.
\newblock Artificial intelligence, machine learning, automation, robotics,
  future of work and future of humanity: A review and research agenda.
\newblock {\em Journal of Database Management (JDM)}, 30(1):61--79, 2019.

\bibitem{abuduweili2019adaptable}
Abulikemu Abuduweili, Siyan Li, and Changliu Liu.
\newblock Adaptable human intention and trajectory prediction for human-robot
  collaboration.
\newblock {\em arXiv preprint arXiv:1909.05089}, 2019.

\bibitem{brose2010role}
Steven~W Brose, Douglas~J Weber, Ben~A Salatin, Garret~G Grindle, Hongwu Wang,
  Juan~J Vazquez, and Rory~A Cooper.
\newblock The role of assistive robotics in the lives of persons with
  disability.
\newblock {\em American Journal of Physical Medicine \& Rehabilitation},
  89(6):509--521, 2010.

\bibitem{pousada2018determining}
Thais Pousada, Betania Groba, Laura Nieto-Riveiro, Alejandro Pazos, Emiliano
  D{\'\i}ez, and Javier Pereira.
\newblock Determining the burden of the family caregivers of people with
  neuromuscular diseases who use a wheelchair.
\newblock {\em Medicine}, 97(24), 2018.

\bibitem{wang2012personal}
Hongwu Wang, Garrett~G Grindle, Jorge Candiotti, Chengshiu Chung, Motoki Shino,
  Elaine Houston, and Rory~A Cooper.
\newblock The personal mobility and manipulation appliance (permma): A robotic
  wheelchair with advanced mobility and manipulation.
\newblock In {\em 2012 Annual International Conference of the IEEE Engineering
  in Medicine and Biology Society}, pages 3324--3327. IEEE, 2012.

\bibitem{yamamoto2018human}
Takashi Yamamoto, Tamaki Nishino, Hideki Kajima, Mitsunori Ohta, and Koichi
  Ikeda.
\newblock Human support robot (hsr).
\newblock In {\em ACM SIGGRAPH 2018 emerging technologies}, pages 1--2. 2018.

\bibitem{wang2020towards}
Zhongdao Wang, Liang Zheng, Yixuan Liu, Yali Li, and Shengjin Wang.
\newblock Towards real-time multi-object tracking.
\newblock In {\em European Conference on Computer Vision}, pages 107--122.
  Springer, 2020.

\bibitem{simonyan2014very}
Karen Simonyan and Andrew Zisserman.
\newblock Very deep convolutional networks for large-scale image recognition.
\newblock {\em arXiv preprint arXiv:1409.1556}, 2014.

\bibitem{he2016deep}
Kaiming He, Xiangyu Zhang, Shaoqing Ren, and Jian Sun.
\newblock Deep residual learning for image recognition.
\newblock In {\em Proceedings of the IEEE conference on computer vision and
  pattern recognition}, pages 770--778, 2016.

\bibitem{howard2017mobilenets}
Andrew~G Howard, Menglong Zhu, Bo~Chen, Dmitry Kalenichenko, Weijun Wang,
  Tobias Weyand, Marco Andreetto, and Hartwig Adam.
\newblock Mobilenets: Efficient convolutional neural networks for mobile vision
  applications.
\newblock {\em arXiv preprint arXiv:1704.04861}, 2017.

\bibitem{girshick2014rich}
Ross Girshick, Jeff Donahue, Trevor Darrell, and Jitendra Malik.
\newblock Rich feature hierarchies for accurate object detection and semantic
  segmentation.
\newblock In {\em Proceedings of the IEEE conference on computer vision and
  pattern recognition}, pages 580--587, 2014.

\bibitem{redmon2016you}
Joseph Redmon, Santosh Divvala, Ross Girshick, and Ali Farhadi.
\newblock You only look once: Unified, real-time object detection.
\newblock In {\em Proceedings of the IEEE conference on computer vision and
  pattern recognition}, pages 779--788, 2016.

\bibitem{redmon2017yolo9000}
Joseph Redmon and Ali Farhadi.
\newblock Yolo9000: better, faster, stronger.
\newblock In {\em Proceedings of the IEEE conference on computer vision and
  pattern recognition}, pages 7263--7271, 2017.

\bibitem{redmon2018yolov3}
Joseph Redmon and Ali Farhadi.
\newblock Yolov3: An incremental improvement.
\newblock {\em arXiv preprint arXiv:1804.02767}, 2018.

\bibitem{bochkovskiy2020yolov4}
Alexey Bochkovskiy, Chien-Yao Wang, and Hong-Yuan~Mark Liao.
\newblock Yolov4: Optimal speed and accuracy of object detection.
\newblock {\em arXiv preprint arXiv:2004.10934}, 2020.

\bibitem{abuduweili2021robust}
Abulikemu Abuduweili and Changliu Liu.
\newblock Robust nonlinear adaptation algorithms for multitask prediction
  networks.
\newblock {\em International Journal of Adaptive Control and Signal
  Processing}, 35(3):314--341, 2021.

\bibitem{abuduweili2023online}
Abulikemu Abuduweili and Changliu Liu.
\newblock Online model adaptation with feedforward compensation.
\newblock In {\em Conference on Robot Learning}, pages 3687--3709. PMLR, 2023.

\bibitem{nassif2019speech}
Ali~Bou Nassif, Ismail Shahin, Imtinan Attili, Mohammad Azzeh, and Khaled
  Shaalan.
\newblock Speech recognition using deep neural networks: A systematic review.
\newblock {\em IEEE access}, 7:19143--19165, 2019.

\bibitem{halageri2015speech}
Akhilesh Halageri, Amrita Bidappa, C~Arjun, Madan~Mukund Sarathy, and Shabana
  Sultana.
\newblock Speech recognition using deep learning.
\newblock {\em International Journal of Computer Science and Information
  Technologies}, 6(3):3206--3209, 2015.

\bibitem{hannun2014deep}
Awni Hannun, Carl Case, Jared Casper, Bryan Catanzaro, Greg Diamos, Erich
  Elsen, Ryan Prenger, Sanjeev Satheesh, Shubho Sengupta, Adam Coates, et~al.
\newblock Deep speech: Scaling up end-to-end speech recognition.
\newblock {\em arXiv preprint arXiv:1412.5567}, 2014.

\bibitem{amodei2016deep}
Dario Amodei, Sundaram Ananthanarayanan, Rishita Anubhai, Jingliang Bai, Eric
  Battenberg, Carl Case, Jared Casper, Bryan Catanzaro, Qiang Cheng, Guoliang
  Chen, et~al.
\newblock Deep speech 2: End-to-end speech recognition in english and mandarin.
\newblock In {\em International conference on machine learning}, pages
  173--182. PMLR, 2016.

\bibitem{lin2014microsoft}
Tsung-Yi Lin, Michael Maire, Serge Belongie, James Hays, Pietro Perona, Deva
  Ramanan, Piotr Doll{\'a}r, and C~Lawrence Zitnick.
\newblock Microsoft coco: Common objects in context.
\newblock In {\em European conference on computer vision}, pages 740--755.
  Springer, 2014.

\end{thebibliography}


\end{document}